\documentclass[twoside,twocolumn]{article}

\usepackage{blindtext} 
\usepackage{graphicx}
\usepackage[sc]{mathpazo} 
\usepackage[T1]{fontenc} 
\usepackage{microtype} 

\usepackage[english]{babel} 

\usepackage[hmarginratio=1:1,right=18mm, left= 18mm, top=25mm,columnsep=35pt]{geometry} 
\usepackage[hang, small,labelfont=bf,up,textfont=it,up]{caption} 
\usepackage{booktabs} 

\usepackage{enumitem} 
\setlist[itemize]{noitemsep} 

\usepackage{abstract} 

\usepackage{titlesec} 
\renewcommand\thesection{\Roman{section}} 
\renewcommand\thesubsection{\roman{subsection}} 
\titleformat{\section}[block]{\large\scshape\centering}{\thesection.}{1em}{} 
\titleformat{\subsection}[block]{\large}{\thesubsection.}{1em}{} 

\usepackage{fancyhdr} 
\usepackage{titling} 

\usepackage{hyperref} 

\pretitle{\begin{center}\Huge\bfseries} 
\posttitle{\end{center}} 
\title{\LARGE Overview of chemical ontologies } 
\author{\textsc{Christian Pachl$^1$, Nils Frank$^1$, Jan Breitbart$^1$ \& Stefan Br\"ase$^2$} } 

\date{} 

\begin{document}

\maketitle
 \footnotetext[1]{Institute of Organic Chemistry, Karlsruhe Institute of Technology (KIT), Fritz-Haber-Weg 6, 76131 Karlsruhe, Germany}
 \footnotetext[2]{Institute of Organic Chemistry, Karlsruhe Institute of Technology (KIT), Fritz-Haber-Weg 6, 76131 Karlsruhe, Germany \\
Institute of Biological and Chemical Systems - Functional Molecular Systems (IBCS-FMS), Karlsruhe Institute of Technology (KIT), Hermann-von-Helmholtz-Platz 1, D-76344 Eggenstein-Leopoldshafen, Germany}


\section{Introduction}
Ontologies order and interconnect knowledge of a certain field in a formal and semantic way so that they are machine-parsable. They try to define allwhere acceptable definition of concepts and objects, classify them, provide properties as well as interconnect them with relations (e.g. "A is a special case of B"). More precisely, Tom Gruber defines Ontologies as a \textit{"specification of a conceptualization; [...] a description (like a formal specification of a program) of the concepts and relationships that can exist for an agent or a community of agents."} \cite{gruber} \\

An Ontology is made of \textit{Individuals} which are organized in \textit{Classes}. Both can have \textit{Attributes} and \textit{Relations} among themselves. Some complex Ontologies define \textit{Restrictions}, \textit{Rules} and \textit{Events} which change attributes or relations. To be computer accessible they are written in certain ontology languages, like the \textit{OBO} language or the more used \textit{Common Algebraic Specification Language}. With the rising of a digitalized, interconnected and globalized world, where common standards have to be found, ontologies are of great interest. \cite{interest}   \linebreak 

Yet, the development of chemical ontologies is in the beginning. Indeed, some interesting basic approaches towards chemical ontologies can be found, but nevertheless they suffer from two main flaws. Firstly, we found that they are mostly only fragmentary completed or are still in an architecture state.  Secondly, apparently no chemical ontology is widespread accepted. Therefore, we herein try to describe the major ontology-developments in the chemical related fields \textit{Ontologies about chemical analytical methods}, \textit{Ontologies about name reactions} and \textit{Ontologies about scientific units}.  \linebreak

Some of the below mentioned Ontologies are licensed under several CC license-types. CC stands for "Creative Commons" copyright licenses which try to establish a more balanced license type than the traditional  "all rights reserved" setting. The CC BY licence \textit{"lets others distribute, remix, adapt, and build upon your work, even commercially, as long as they credit you for the original creation."}. The CC BY-SA license adds the restriction that ones own creations are licensed \textit{"under the identical terms. [...] All new works based on yours will carry the same license."} \cite{cc}

\section{Ontologies about chemical analytical methods}
Beside several simple wikipedia lists (under CC BY-SA 2.5) \cite{wiki1} \cite{wiki2} \cite{wiki3} \cite{wiki4} \cite{wiki5}  which do not quite fulfil the above mentioned characteristics of an ontology, four ontologies are of further interest: \textit{CHMO}, \textit{The ChAMP Project}, \textit{AnIML Technique definition} and \textit{Allotrope}. \linebreak

\begin{figure}
\includegraphics[width=0.9\linewidth]{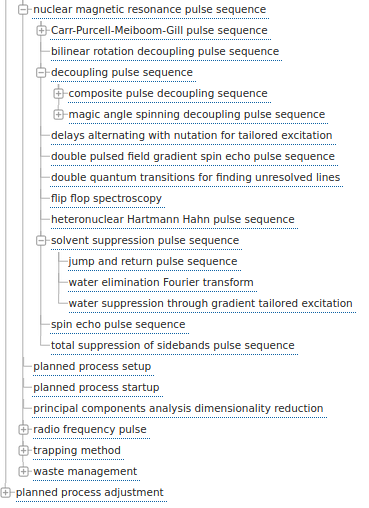}
\caption{Exemplary branch of the \textit{CHMO} (derived by  \cite{screenshotCHMO})} \label{screenshotCHMO}
\end{figure}

The \textit{Chemical Method Ontology}, abbreviated with \textit{CHMO} \cite{chmo}, uses the definitions out of the established \textit{IUPAC Orange Book} \cite{iupacOrange} and convert them into an ontology language (OBO and OWL). Thus the \textit{CHMO} contains several hundreds analytical methods, classified in method-families, described and equipped with synonyms (see Figure \ref{screenshotCHMO}). Despite this broad scope no further metadata for each analytical method is provided. The CHMO is lincensed under CC BY 4.0 \cite{chmo} and is actively updated. \\

The \textit{Allotrope} Ontology \cite{allotrope} takes over parts of CHMO and can be therefore be seen as redundant towards the \textit{CHMO}. \\

The \textit{The ChAMP Project}  \cite{champ} provides the design for a possible ontology structure without actually including concrete chemical analytical methods. It appears as a project with high ambition, but currently it is not under active development any more (February 2020). \pagebreak

Finally, the \textit{AnIML Technique definition} ontology \cite{animl} provides the ontology-realization of a certain area of chemical analytical method: UV-vis spectroscopy and chromatography methods. Therefore, it describes not the whole field of chemical methods but rather gives an in-depth ontology on UV-vis and chromatography methods with rich metadata (e.g. temperature, density of the sample, Center frequency, Spectral Post-Processing, ...). The integration of  IR-, MS-, NMR-spectrocopy was planned, but the project appears to be currently inactive (February 2020). No license was found. \\

To conclude, the \textit{CHMO} describes the landscape of chemical analytical methods but suffers from missing metadata despite a well-made hierarchical structure. The missing metadata limits its possible use in the implementation into lab-software but can serve as a sufficient good starting point for the development of an adapted analytical ontology (adding own content via editing-software like \textit{Stanford Prot\'{e}g\'{e}} \cite{protege}). In contrast, the \textit{AnIML Technique definition} ontology is incomplete but is a good example how metadata could support the processing of different analytical data into a coherent machine-readable system in chemical analytical facilities. A further summary is given in Table 1.

\begin{table}[h]\captionsetup{singlelinecheck=true} 
 \caption{Summary of Chemical Analytical Method Ontologies}\def\arraystretch{1.25}  \centering  \hspace*{-0.5cm} \begin{tabular}{cp{4.5cm}c} 
\toprule
 & Characteristics & License \\ \midrule
CHMO & broad scope, nearly complete, but no further metadata & CC BY 4.0 \\
Allotrope & Takes over parts of CHMO & not available \\
ChAMP & proposal for analytical ontology structure, no content (project asleep) & not available \\
AnIML & focus on UV-vis and chromatography with rich metadata & not available \\
 \bottomrule
\midrule
\end{tabular}
\end{table}

\section{Ontologies about chemical reactions}
Name reactions serve a key purpose in chemistry as they provide a certain key-word for a given reaction, it's educts and products as well as data about the reaction environment. Therefore an ontology, classifying these reactions is of high interest in order to be able to properly suggest reaction paths for synthetical chemistry. \\

Similar to analytical methods, there are very few ontologies to be found for this specific field. While there are again Wikipedia lists of named reactions \cite{wiki6} \cite{wiki7}; these are again machine-unparsable. Also they do not provide any sorting. Furthermore, \cite{wiki6} mixes organic and inorganic name reactions. There are however a few projects to be found to fulfil the characteristics of an ontology: The \textit{RXNO}, \textit{MOP} and \textit{KEGG reaction database}. In addition, some smaller or unfinished projects such as \textit{SWEET} and \textit{PIERO} are available.\\

\begin{figure}
\includegraphics[width=0.8\linewidth]{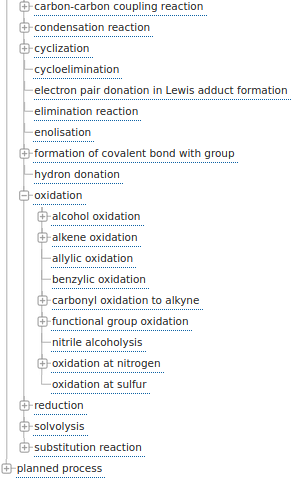}
\caption{Exemplary branch of the \textit{RXNO} (derived by  \cite{screenshot})} \label{screenshotRXNO}
\end{figure}

The \textit{RXNO} \cite{RXNO} is by far the most complete ontology out of the ones here presented. It contains over 500 name reactions, sorted into an ontology with several layers. In the first instance, the reactions are sorted by the general type of the reaction, e.g. oxidations or cyclizations. The second layer is dividing the reactions further into smaller categories, for example their dedicated reactants. Again, using the example of oxidations, they are further divided into reactions describing the synthesis of alcohols or alkenes (see Figure \ref{screenshotRXNO}). Using this tree structure, a reaction can obtained that is explicitly designed for a given goal. However, despite the enormous amount of reactions, sorted into this system, there is no further information given about these, aside from their name and parents. Also, certain chemicals, which are needed for the reaction, are assigned to the process and vice versa. There is no licence given. \\

Similar to RXNO, \textit{MOP} is a very extensively-evolved	 ontology \cite{MOP}. It is very similar to RXNO's structure, however, not identical. While the RXNO lists actual name reactions in organic chemistry, MOP only lists reaction mechanisms of general types.  It only contains the branches of RXNO under "molecular process" omitting concrete name reactions. This ontology is perspicously sorted and actively updated, while no licence was found. Both, RXNO and MOP have been published by the same person, Colin Bachelor. \cite{MOP github}\\

\textit{KEGG reaction database} is a database for enzymatic reactions \cite{KEGG}. It is divided into two parts, reaction module and reaction class. Reaction class sorts hundreds of reactions very accurately in several sub-layers. The reactions themselves contain a lot of metadata, such as name, synonymes, substrate, products, references, and a description of the reaction. In contrast to the reaction class, which is sorted by informations of certain enzymes being involved, the reaction module assorting reactions only based on their chemical identity. This makes it similar to RXNO and MOP, while KEGG still only handles reactions that are relevant for biochemical reactions. Moreover, no actual defined names of the described enzymatic reactions are given. This part is not as strictly sorted, there is a list of certain reaction goals which lead to a broad list of applications of the respective reaction type. There aren't any subclasses. \textit{KEGG reaction database} does not fulfil the requirements for an ontology as it is not saved in a machine parsable language. However, it is a very large database for reactions. This database is activly updated. If one wants to use this database, there is a request form linked on their website \cite{KEGG request}. \\

\textit{SWEET} is an ontology attempt similar to RXNO or MOP \cite{SWEET}. However this ontology so far is an empty shell, there is only the first layer of an hierarchy examined. Considering the website, it was ment to be a very big ontology, however this project appears to be inactive. It is licensed under CC0 1.0 \cite{SWEET copyright}. \\

Lastly, the \textit{PIERO}-project \cite{PIERO} was an attempt to sort and extend the \textit{KEGG}-reactions into a full ontology. However, this project gives no further information about. Actually, it has been last updated in February 2015.\\

It has to be mentioned, that there are several attempts and projects that use reaction ontologies to predict reaction paths. Some of them are inactive, such as the \textit{EROS} project \cite{EROS}, while there are also published applications such as the \textit{Reaxys synthesis planner} \cite{Reaxys}. There is however no access to the raw data behind these programms as they are proprietary. \\

To conclude this examination of reaction ontologies: There are currently two up-to date and usable ontologies, \textit{RXNO} and \textit{MOP}. Both of them are usable and fulfil the requirements defined at the beginning. Both do not contain a lot of metadata, which limits their current application in lab-software and other usage cases. Despite this fact, these ontologies can serve as a great starting point for further projects. There is the \textit{KEGG reaction database}, which, while not being saved in an OBL or OWL format contains a great amount of information, which possibly could be integrated in some sort of software. The main problem with this database, besides its unfortunately saved data, is the fact that its copyright terms are not publicly announced. Therefore, a usage is questionable. There are several currently inactive projects like \textit{PIERO} and \textit{SWEET} as well as Wikipedia lists, which can serve as a starting point for further research on reactions. There are also programs that predict reactions paths, such as the Reaxys-synthesis planner. A  summary is given in Table 2. \\

\begin{table}[h]\captionsetup{singlelinecheck=true}
 \caption{Summary of Chemical Reactions Ontologies}\def\arraystretch{1.25} \centering \hspace*{-0.5cm} \begin{tabular}{cp{4.5cm}c}
\toprule
 & Characteristics & License \\ \midrule
RXNO & mostly completed ($>$ 500 specific name reactions), no further metadata, organized in simple tree structure & not available \\
MOP & General reaction types & not available \\
KEGG & Enzymatic reactions sorted by reaction type as well as enzymatic class. Rich metadata available. & By permission \\
SWEET & Alternative to RXNO. Empty project, asleep & CC0 1.0 \\
PIERO & Extension to KEGG. Asleep & not available \\
 \bottomrule
\midrule
\end{tabular}
\end{table}

\section{Ontologies about units}
The implementation of units and their definition is vital for the understanding and communication of data. In order to be able to compare information, it is important to use the same units. Defining units in order to serve better communication worldwide is the goal of the following projects. It is mentionable, that all finished ontologies contain all relevant units and scales due to their arguably limited number. There are units, that are not defined in these, such as the ones in the Wikipedia's "List of humourous units of measurement" \cite{wiki11}, which have no actual application in a scientific field.\\

It is again worth noting that there are several free online accesible definitions for units, which are most handy to use if one needs to look up a specific one or some historical unit \cite{wiki8} \cite{wiki9} \cite{wiki10}. However, these are not applicable for hirarchy-based systems, as they only sort their entries by alphabet. There are ontologies though, that do exactly that:\\

\textit{UO}, the "units ontology" \cite{UO} is considerable to be one of the biggest regarding that. It contains 428 sorted and machine-parsable units, including their definition and synonyms. It is sorted in a three-level system, each layer being sorted alphabetical. This ontology is designed to be applicable in a broad field of sytems. It is constantly updated and is covered by the CC-BY 4.0 licence \cite{UO copyright}.\\

\textit{OM 2} \cite{OM 2}: The "Ontology of units of Measure" is an ontology based on several official papers by NIST, such as the "Guide for the Use of International System of Units (SI)" \cite{NIST}. The supposed fields of applications are, according to the authors \cite{OM 2 app}, thermodynamics, economics, chemistry and many more. It contains a wide range of units, sorted into a multi layered system, however in a rather unstrigent layout. It has to be said that this ontology lacks of metadata, like the  acceptance of use of a cerain unit. However, it contains a description, synonyms and the definition in SI-units. It is saved in the OWL language and is licensed by CC-BY 2.0.\cite{OM 2 copyright}\\

\textit{OM 1.8} \cite{OM 1.8} is an outdated version of the previously descibed \textit{OM 2} ontology. It contains similar data, has the same design and structure and is protected by the same copyright.\\

\textit{QUDT} \cite{QUDT} is an initiative that gathered information about various topics, one of them are units. It is a set of vocabulary related to the give topic, however, written in the protege readable format OWL \cite{QUDT units}. It includes the information about the name, symbol, description as well as a label for each unit. The fact, that this ontology is only defining vocabularies, makes already clear that it does not contain any further structure in which the vocabulary is sorted in. There are similar set of vocabulary for constants, quantities, scales, etc. on their github page \cite{QUDT github}. It is a potentional interesting ontology, as it seems to be actively updated. It falls under CC-BY 4.0 copyright.\\

Furthermore, there have been initiatives for designing unit-ontologies in the past, for example the \textit{OASIS-QUOMOS}-project \cite{OASIS}. There is no final repository to be found, as this project was abandoned in August 2014.\\

For the field of unit ontologies, there are more sucessfull projects to be found than for the other chemical-related topics. As all of them are in terms of contained information completed, there does not seem to be a major difference between \textit{UO}, \textit{OM 2} and \textit{QUDT}. In terms of clarity and structure of the project, the first one is to recommend. There are also some outdated ontologies which are ommited here. A further summary of active ontologies is given in Table 3.

\begin{table}[h]\captionsetup{singlelinecheck=true}
 \caption{Summary of Unit Ontologies}\def\arraystretch{1.25} \hspace*{-0.5cm} \centering \begin{tabular}{cp{4.5cm}c}
\toprule
 & Characteristics & License \\ \midrule
UO & biggest unit-ontology; sorted in three-level system - within them alphabetically & CC-BY 4.0 \\
OM 2 & Based on NIST-specifications; unstringently sorted; some metadata & CC-BY 2.0 \\
OM 1.8 & Prior version of OM 2 & CC-BY 2.0 \\
QUDT & Defines vocabulary about units & CC-BY 4.0 \\
OASIS & Asleep & not available \\
 \bottomrule
\midrule
\end{tabular}
\end{table}

\section{Summary and Conclusion}

Ontologies are subject of current initiatives and scientific research. They are vital to digitalize chemistry to make it more connective and clear. Also, they serve the purpose of enhancing the communication across various countries and languages. However, some of these ontologies include immense amout of data, which makes them really hard to grasp and keep them up to date. Considering this, it is no surprise that we found many inactive or unfinished projects. This confirms the importance of FAIR-data (findable, acessable, interoperable, re-usable) \cite{FAIR} in the so-called field of Cheminformatics. There are currently various active ontologies found on repositories such as the "Ontology Lookup Service" \cite{OLS}. Most Ontologies to be found are also free-acessible by the public and free to re-use. As ontologies are a current subject of interest, it is to expected, that the results presented in this article may be completely outdated in a few years.


\end{document}